# Large Language Model probabilities cannot reliably distinguish between possible and impossible language

Evelina Leivada[1,2*], Raquel Montero[1], Paolo Morosi[1*], Natalia Moskvina[1], Tamara Serrano[1], Nikoleta Pantelidou[1], Marcel Aguilar[3] & Fritz Günther[4]

[1] Departament de Filologia Catalana, Centre de Lingüística Teòrica, Universitat Autònoma de Barcelona, Barcelona 08193, Spain

[2] Institució Catalana de Recerca i Estudis Avançats (ICREA), Barcelona, 08010, Spain

[3] Similarweb, Spain

[4] Institut für Psychologie, Humboldt-Universitat zu Berlin, Berlin 10099, Germany

* Corresponding authors

E-mail: evelina.leivada@uab.cat (EL)

E-mail: paolo.morosi@uab.cat (PM)

# Abstract

A controversial test for Large Language Models concerns the ability to discern possible from impossible language. While some evidence attests to the models' sensitivity to what crosses the limits of grammatically impossible language, this evidence has been contested on the grounds of the soundness of the testing material. We use model-internal representations as a direct means of investigating the way Large Language Models represent the 'grammatical-ungrammatical' distinction. In a novel benchmark, we elicit sum log probabilities from 8 models and compute minimal-pair surprisal differences, juxtaposing probabilities assigned to grammatical sentences to probabilities assigned to (i) lower frequency grammatical sentences, (ii) ungrammatical sentences, (iii) semantically odd sentences, and (iv) pragmatically odd sentences. The prediction is that if string-probabilities can function as proxies for the limits of grammar, the ungrammatical condition will stand out among the conditions that involve linguistic violations, showing a spike in the surprisal rates. Our results do not reveal a unique surprisal signature for ungrammatical prompts: across models, we observe comparable surprisal for the pragmatically odd condition, and the semantically odd condition consistently shows even higher surprisal. We thus demonstrate that probabilities do not constitute reliable proxies for grammaticality, showing that probability differences alone cannot be used to infer model-internal representations of syntactic knowledge. Consequently, claims about models being able to distinguish possible from impossible language need verification through a different methodology.

# Introduction

Large Language Models (LLMs) are Artificial Intelligence (AI) systems designed to interact using human language. These models produce synthetic language that looks remarkably like human-produced language, after being trained on vast datasets consisting of text, images, and/or speech. By employing deep learning techniques, LLMs can perform a wide range of tasks, including asking and answering questions, summarizing content, and translating content into different languages. Their ability to engage in written communication makes them good tools with applied uses in a variety of contexts, while their remarkably human-like behavior has led to the claim that the current generation of LLMs has successfully passed the Turing test [1-2]; a test of an artificial agent's ability to exhibit intelligent behavior that can be indistinguishable from that of a human in a given conversational context. If the judge cannot reliably distinguish the identity of the interlocutor (i.e. artificial agent or human), the artificial agent is said to have passed the test.

In this context, the linguistic abilities of LLMs, and specifically the degree to which they deviate from human baselines in linguistic tests, either quantitatively, in terms of percentage of errors, or qualitatively, in terms of kinds of errors and whether they are human-like or not, is currently a matter of debate. On the one hand, some experiments find that LLMs perform worse than humans in tasks that ask whether a particular structure is well-formed and/or appropriate in a target natural language: Models may lag behind humans in either accuracy (i.e. providing the target answer) or stability (i.e. providing the target answer consistently) [3-8]. Crucially, in this testing regime that involves eliciting judgments of well-formedness, models are asked to evaluate the soundness of linguistic prompts in a natural language question-answering scenario that requires inferential operations to construe the interpretation of the given prompts. Humans excel in this *prompting* task, thus offering a solid baseline for evaluating whether the linguistic competence of LLMs is indeed human-like [4]. On the other hand, the appropriateness of running such tests to determine the language abilities of LLMs has been questioned on the grounds that this can be viewed as a metalinguistic task, thus not fully appropriate for models [9]. At their core, LLMs are complex network architectures trained with the objective to estimate the probability of a token given a context; consequently, they estimate probabilities for each word in a text, and by extension produce probabilities for texts, including sentences. As an alternative, that line of work therefore suggests that *direct probability measurements* for given linguistic expressions can be more informative than prompting and observing models' verbal output, when the aim is to determine whether a model has formed a linguistic generalization of interest [9-11]. Indeed, measuring the probabilities that models assign to different strings of words has provided results suggesting that LLMs, and possibly even their precursors (RNNs), assign higher probabilities (or lower surprisal = -log(prob)) to grammatical sentences than ungrammatical sentences in minimal pairs. This probability difference between well-formed from ill-formed prompts has been interpreted as showing that these models are able to acquire deeper grammatical competence and to form linguistic generalizations [9-14].

Undoubtedly, adopting the right testing methods is important in order to correctly determine the outcome of the "acquisition" process in LLMs [10, 15-16], not the least because their ability to tell apart grammatically possible from impossible prompts has been recently challenged on theoretical grounds by Chomsky et al. (2023; see also Moro et al. 2023 for a similar yet not identical claim about how the distinction between possible versus impossible languages cannot be precisely formulated and fed to LLMs) [17-18]. The terms *possible* and *impossible* are to be understood in their linguistic sense. A *possible* linguistic expression is one that conforms to the grammatical constraints of the language under investigation, whereas an *impossible* expression violates those constraints and therefore cannot constitute a well-formed expression of that language. Importantly, linguistic impossibility is distinct from semantic or pragmatic anomaly: an expression may be grammatically possible while being difficult to interpret semantically or infelicitous in a particular context.

Albeit theoretically plausible, the alleged inability of LLMs to distinguish possible from impossible languages remains to be tested and proven. From a theoretical perspective, the stakes are high: Any evidence for or against this claim bears great significance, because this point represents the crux of the matter in the battle of frameworks within linguistics. Perhaps the best example of this clash can be found in Piantadosi (2023) [19]. Commenting on how LLMs can learn morphosyntax through forming generalizations based solely on statistical relations between words —a point that Piantadosi argues has been contested by some generative linguists (cf. Adger's claim that there is no current theory of linguistics that reduces the *human* ability to figure out syntax to general probabilistic relations between elements of unanalysed data)[20]— Piantadosi argues that this is exactly what the current generation of LLMs does. Leaving aside the fact that Adger's claim focuses on how *humans* —not LLMs— learn language, the core of Piantadosi's claim remains to be empirically verified. Do LLMs form linguistic generalizations that enable them to discern grammatically possible from impossible languages, and how do we evaluate this? If prompting is not the right task, can probabilities offer a more reliable window into the inner linguistic understanding of LLMs?

Aiming to adjudicate this point, in pioneering work, Kallini et al. (2024) find that GPT-2 small models struggle to learn impossible languages when compared to English. Concretely, they find that GPT-2, after training, showed higher perplexity scores (i.e. a metric that evaluates how well the model's predicted probability distribution for the next word matches the actual next word in a string of words) in impossible structures compared to their control grammatical counterparts [14]. Hunter (2025) notes a confound in Kallini et al.'s design [21]. When comparing their possible (and minimally deviating from English) NoHop condition (e.g., "He clean **s** his very messy bookshelf") and their impossible WordHop condition (e.g., "He clean his very messy bookshelf **s**)", Hunter argues that a confound exists between the application of a rule that is count-based and a rule that creates non-adjacent dependencies. Indeed, count-based rules such as 'place the negation particle in the third position after the verb' are thought to create unreal, linearly based relationships that are not processed by the human brain as their licit, constituency-based counterparts [22]. Thus, the issue according to Hunter is that Kallini et al. do not compare constituency-based non-adjacent dependencies (which the human brain can naturally accommodate and parse) to count-based non-adjacent dependencies, but *adjacent* dependencies to count-based non-adjacent dependencies. However, as Hunter also recognizes, this still leaves the main issue unsettled, as his re-analysis of Kallini et al.'s stimuli is not informative about whether LLMs truly exhibit a human-like asymmetry between count-based and constituency-based rules [21]; an asymmetry that would entail sensitivity as to what crosses the limits of possible language.

Yang et al. (2025) reach a somewhat similar conclusion to Kallini et al., testing a variety of languages [23]. However, they also note that while GPT-2 tends to learn possible (i.e. attested) languages better than impossible ones, its perplexity scores do not always distinguish between the two types of languages. For example, impossible Italian yields a numerically lower perplexity than natural Italian. These inconsistencies cast some doubt on the claim that LLMs can *reliably* tell possible from impossible languages.

In other recent work, Hu et al. (2024) argue that probabilities elicited over minimal-pair prompts (i.e. prompts that are identical apart from one difference) show that LLMs have learned the distinction between the possible and the impossible: The less surprising (i.e., the more probable) a sentence relative to its minimal-pair counterpart is, the more likely models (and humans) are to judge it as grammatical [11]. In other words, surprisal values in LLMs align with well-formedness judgments in humans in the following way: The grammatical sentence in a 'grammatical-ungrammatical' pair is the one that both humans and models "prefer": humans give it a higher rate of acceptability, while models find it less surprising = more probable than its counterpart. This finding has given rise to the claim that obtaining direct probability estimates through fetching surprisal values is a good proxy for grammaticality [11].

In sum, the prediction that arises from the previous literature is that a correlation between probability assignment and grammaticality exists: Hu et al. (2024) argue that a controlled manipulation that consists of probability measurements over minimal pairs "*isolates grammatical differences*, so a model that has learned the correct generalizations should assign higher probability to the grammatical sentence in each minimal pair". In more recent work, Hu et al. (2026) modify the strength of this correlation, somewhat toning it down, by suggesting that other factors (e.g., message probability) can overwhelm the contribution of grammaticality in determining probability [24]. At the same time, the core of the argument is maintained in their conclusions; according to Hu et al. (2026), critical information about a model's underlying grammatical knowledge can be inferred by inspecting string probabilities [24]. Similarly, Kauf et al. (2024) find that model probabilities reliably reflect event knowledge in LLMs [25]. The tested models tend to assign a higher likelihood to possible vs. impossible events (e.g., "The teacher bought the laptop" vs. "The laptop bought the teacher"), although they are less consistent in distinguishing likely vs. unlikely events (e.g., "The nanny tutored the boy" vs. "The boy tutored the nanny") [25].

Observing that probabilities are used as proxies for *semantico-pragmatic knowledge* in Kauf et al. (2024), where the focus is on events that are (im)possible and/or (un)likely in the real world, while they are used as proxies for *syntactic knowledge* in Hu et al. (2024, 2026), where higher probabilities are explicitly linked to grammatical generalization capabilities, begs the question of what kind(s) of model-internal knowledge do probabilities eventually represent. This is the question addressed in the present work. Can we causally infer grammaticality by measuring the probabilities models assign to strings of words? Are probability differences a specific marker of grammaticality? Put another way, can probabilities provide a window into whether LLMs are sensitive to grammatically possible vs. impossible prompts?

This is still an open matter for various reasons. First, while surprisal values indeed reflect grammatical well-formedness, they are also presumably sensitive to many other influences such as word frequency, ambiguity, idiomaticity, collocability, semantico-pragmatic soundness, etc. Second, when it comes to figuring out the syntactic abilities of LLMs, previous research, that has argued for the superiority of the probability measurement method over prompting, has largely used 'grammatical-ungrammatical' pairs. This means that the experimental design has *restricted* the source of surprisal to one possible cause: ungrammaticality. To illustrate the problem with an analogy, arguing that probabilities are a good proxy for grammaticality —given that ungrammaticality spikes the surprisal rates— amounts to arguing that facial expressions are a good proxy for determining excessive salt in a dish: While too much salt will likely give rise to an expression of disgust, we cannot infer [+excessive salt] from the latter, because disgust can occur in the absence of excessive salt (e.g., due to excessive oregano). In an experimental setting that exclusively features [+/-excessive salt] dishes, the difference in facial expressions indeed will align with quantity of salt. However, this does not legitimize the conclusion that [+/-excessive salt] can be *inferred* from facial expressions, unless one establishes that one specific expression of disgust is unique to the property in question: [+excessive salt]. Likewise, even if ungrammaticality tends to increase surprisal, surprisal may also increase in response to semantic or pragmatic deviations. The fact that these sources of surprisal can overlap in terms of their numerical expression makes it difficult to infer grammaticality from surprisal alone.

We thus identify the following gap in the literature: To what extent does grammaticality have a unique surprisal signature that enables us to legitimately treat probabilities as proxies for grammaticality? The Research Question is formulated as follows: Does the 'grammaticality-ungrammaticality' distinction have a unique surprisal signature, such that we can infer ungrammaticality based on observing minimal-pair surprisal differences?

# Method

We use internal model representations to directly investigate the internal states of the tested models. In a novel benchmark, we elicit probabilities from 8 models that come from different families: Gemma-2b and Gemma-7b, Pythia-1b-deduped, Mistral-7b, DeepSeek-R1-Distill-Qwen-32b, Olmo-3.1-32b-Instruct, Qwen-2.5-32b-Instruct and Qwen-2.5-7B-Instruct [26-31]. The task taps into minimal-pair surprisal differences, juxtaposing 50 grammatical sentences with (i) 50 lower frequency grammatical sentences, (ii) 50 ungrammatical sentences, (iii) 50 pragmatically odd sentences, and (iv) 50 semantically odd sentences. Our design is based on Hu et al.'s (2024) and Kallini et al.'s (2024) rationale that measuring the probabilities models assign to strings in a minimal-pair settings has the potential to reveal that LLMs possess strong, human-like grammatical generalization capabilities [11, 14]. In this context the hypothesis we seek to test is whether probabilities can be construed as a proxy for ungrammaticality. If this is the case, the prediction that follows is that the ungrammatical condition will stand out among the 3 tested conditions that involve violations, showing a spike in the surprisal rates. According to surprisal theory, the processing difficulty of a word is proportional to its surprisal (i.e. its negative log-probability) in the context within which it appears, thus predicting a linear relation between surprisal and string processing difficulty [32-33], formalized as follows: $S(w_i) = -\log_2 P(w_i|w_1, w_2 \ldots w_{i-1})$. Table 1 summarizes the tested conditions, offering examples and providing predictions regarding surprisal.

**Table 1. Predictions and examples per condition.**

| Condition (C) | Prediction for minimal-pair differences | Examples |
|---|---|---|
| C1: Grammatical (high frequency; HF)-Grammatical (low frequency; LF) | Low | 1a. John was craving a glass of **cold water**. [HF]<br>1b. John was craving a glass of **cold beer**. [LF]<br><br>2a. Last week we found out that Dan was gifted a **luxurious car**. [HF]<br>2b. Last week we found out that Dan was gifted a **luxurious yacht**. [LF] |
| C2: $\text{Grammatical}_{HF}$-Ungrammatical | High | 1c. John was craving a glass of **cold water**.<br>1d. John was craving a glass of **water cold**.<br><br>2c. Last week we found out that Dan was gifted a **luxurious car**.<br>2d. Last week we found out that Dan was gifted a **car luxurious**. |
| C3: Grammatical$_{HF}$-Pragmatically odd | Low | 1e. John was craving a glass of **cold water**.<br>1f. John was craving a glass of **cold tears**.<br><br>2e. Last week we found out that Dan was gifted a **luxurious car**.<br>2f. Last week we found out that Dan was gifted a **luxurious denture**. |
| C4: Grammatical$_{HF}$-Semantically odd | Medium | 1g. John was craving a glass of **cold water**.<br>1h. John was craving a glass of **cold laptops**.<br><br>2g. Last week we found out that Dan was gifted a **luxurious car**.<br>2h. Last week we found out that Dan was gifted a **luxurious toothache**. |

To explain the stimuli, C1 consists of minimal pairs (1a-b, 2a-b, Table 1) of fully grammatical, semantically meaningful, and pragmatically felicitous sentences. The only difference between the two parts of each minimal pair is the frequency of the last two words, with the first part of each pair (1a, 2a) always featuring the more frequent adjective-noun combination (based on the phrasal frequency in the Corpus of Contemporary American English; COCA) [34]. The prediction is that the differences between the two parts will be minimal, given that both prompts are well-formed. Throughout all conditions, the first part of each minimal pair is the same as in C1 to ensure comparability.

The ungrammatical sentences of C2 fall under the Reverse category of impossible languages of Kallini et al. (2024) [14]. In the second part of each pair (1d, 2d), the order of the prefinal and the final tokens is reversed resulting in a sentence that is quite high on the scale of impossible languages according to the Kallini et al. hierarchy of impossibility. If surprisal is a window into the grammatical abilities of LLMs, the prediction is that this will be the condition that will show the biggest difference between the two parts.

C3 compares the high frequency grammatical prompts of C1 with prompts that are pragmatically odd. No rules of grammar are violated in C3, and the semantic requirements of the verbs are satisfied. To illustrate this with an example from Table 1, taking (1f) as reference, while tears are not considered a typical or intended source of nutrition for humans, lachryphagy is attested in other species, and tears are in principle drinkable. This makes (1f) in C3 different from (1h) in C4: while it is pragmatically odd to crave tears, tears are drinkable, but laptops are not. The prompts in C4 do not satisfy typical lexical semantic requirements, which is why we call them semantically odd.

Overall, the prediction across conditions is as follows: C1<C3<C4<C2. This ordering reflects the expected contribution of different sources of violations, rather than a strict claim that grammaticality itself is gradient. C1 should yield the smallest minimal-pair difference because the prompt is grammatical, semantically coherent, and pragmatically felicitous; the only manipulation concerns the frequency of the final lexical combination. C3 should yield a higher surprisal rate than C1 because, although the sentence remains grammatical and semantically interpretable, it contains an atypical adjective+noun combination that is pragmatically odd given real-world contingencies. C4 should yield a still higher surprisal rate because it violates the lexical-semantic selectional requirements of the verb. Hence, although the sentence is syntactically well formed, it is difficult to parse coherently, as laptops cannot be drunk. Finally, C2 is predicted to yield the highest surprisal rate because the reversal of word order directly violates a syntactic requirement, disrupting the grammaticality of the sentence. Thus, the hierarchy reflects a prediction about how strongly each type of deviation should affect string probability. In the two edges of this continuum, we find C1 (no violation whatsoever), and C2, which is expected to show the greatest differences in terms of surprisal, given that the Reverse prompts are highly impossible [14]. Fig 1 illustrates the study design and ranks the conditions in terms of predicted surprisal. The testing stimuli, the dataset, and the code used to run the analysis are available at https://osf.io/pgqab/overview?view_only=f60ff2d09fc4485fb416b3250f7ba265.

[INSERT FIG 1 HERE]

**Fig 1.** Study design and stimuli. The grammatical condition is high on the possibility hierarchy, and every condition below it increases the impossibility, reflecting a decrease in the parsability of the prompts. To explain this, the pragmatically odd condition is easy to parse, as it does not involve any violation, even though it does not comply with typical pragmatic requirements. The other two conditions involve violations that differ in terms of how easily their prompts can be parsed. Adding a second violation in both conditions illustrates this point: ‘John was craving a [colorless green] [glass of cold laptops]’ involves two consecutive “impossible concepts”, but it can be still parsed, and a meaning

can be construed. This is in stark contrast to 'John was craving a [of glass] [water cold]' which involves two consecutive reversals. Crucially, at the onset of the first reversal, the prompt becomes unintelligible as a syntactic unit due to the violation of word-order requirements, and if read out loud, it will be read as a list of individual words, and not as a sentence. This is why C2 prompts are deemed to be more impossible (i.e. understood as more difficult to parse as a whole) than C4 prompts, giving rise to C1<C3<C4<C2.

# Results

First, we plot and analyze surprisal differences for each model. Fig 2 shows surprisal values across models. Fig 3 shows surprisal *differences* to the LF-grammatical condition (C1), the shared reference level for all our conditions. To analyze the data, we employed a linear mixed-effects model predicting surprisal difference by condition (indicated as phenomenon below), word frequency differences and length differences for the critical nouns that differ across conditions, with a random intercept for the test item; in lme4 [35-36] syntax:

diff_surprisal ~ phenomenon + diff_freq + diff_length + (1 | test_item)

The reference level for the "condition" factor was set to C1 (comparison between the HF-grammatical and LF-grammatical condition). The Intercept in the model thus reflects that condition: If the Intercept parameter is positive and significant, this means that the second sentence in the minimal pair has higher surprisal than the control sentence in C1. This is not the case for any LLM: The Intercept is never significantly different from zero (see Table 2). Thus, when comparing HF-grammatical and LF-grammatical prompts, simple frequency differences in the final noun of the sentence do not significantly affect model surprisal.

**Table 2. Intercept estimates for linear mixed models with C1 as the baseline condition. Intercepts significantly different from zero would indicate differences in surprisal between the HF-grammatical and LF-grammatical conditions, which we do not observe.**

| LLM | Intercept: *b* [.95-CI] | *t (df)* | Cohen's *d* | *p* |
|---|---|---|---|---|
| DeepSeek-R1-Distill-Qwen-32B | 0.836 [-0.736, 2.408] | t(117.7) = 1.04 | 0.192 | 0.299 |
| gemma-2b | 0.010 [-1.288, 1.307] | t(108.7) = 0.02 | 0.003 | 0.988 |
| gemma-7b | 0.103 [-1.190, 1.395] | t(115.3) = 0.16 | 0.029 | 0.877 |
| Mistral-7B-v0.1 | 0.002 [-1.379, 1.384] | t(111.2) = 0.00 | 0.001 | 0.997 |
| Olmo-3.1-32B-Instruct | 0.059 [-1.731, 1.850] | t(110.8) = 0.06 | 0.012 | 0.948 |
| pythia-1b-deduped | -0.003 [-1.361, 1.356] | t(115.9) = -0.00 | -0.001 | 0.997 |
| Qwen2.5-32B-Instruct | 0.965 [-0.521, 2.451] | t(114.6) = 1.27 | 0.238 | 0.206 |
| Qwen2.5-7B-Instruct | 1.013 [-0.522, 2.549] | t(117.0) = 1.29 | 0.239 | 0.198 |

If a parameter estimate of the "condition" factor is positive and significant, this means that surprisal differences in the respective condition are larger than in this Intercept condition. This is the case for all conditions across all models (all *b*s > 7.87, all *t*s > 10.93, all *d*s > 1.80, all *p*s < 0.001). Based on this result, we can argue that the tested models are more surprised by prompts that involve linguistic violations than fully well-formed prompts with no violation. At the same time, as Fig 2 shows, there is no clear cut-off point for ungrammaticality [37]. We observe a great overlap in surprisal values across the 3 conditions that involve violations, and even among the conditions that involve no violations and those that do. If minimal-pair comparisons could isolate grammaticality from other factors that may influence surprisal and thus act as proxy for models' sensitivity to grammatically impossible language, we would expect the ungrammatical condition in Fig 2 to be linked to surprisal values that cluster separately from those of the other conditions.

[INSERT FIG 2 HERE]

**Fig 2.** Surprisal scores across models and conditions. HF means High Frequency. LF means Low Frequency.

If we set C2, the ungrammatical condition, as the reference condition comparing all the other conditions to it, we find that while it consistently has higher minimal-pair surprisal differences than C1, the LF-grammatical condition, ($p < 0.001$ for all models), for most LLMs it does not significantly differ from C3, the pragmatically odd condition (see Table 3). The only exception is DeepSeek-R1-Distill-Qwen-32B, where the surprisal differences between the LF-grammatical condition and the pragmatically odd condition are significantly *higher*.

**Table 3. Differences between the C2 (HF-grammatical vs. ungrammatical) and C3 (HF-grammatical vs. pragmatically odd) condition, estimated as the regression weights of the C3 condition in a linear mixed model in which the C2 condition is the baseline condition.**

| LLM | prag. odd C3: *b* [95% CI] | *t (df)* | Cohen's *d* | *p* |
|---|---|---|---|---|
| DeepSeek-R1-Distill-Qwen-32B | 2.250 [0.617, 3.883] | 2.70 (153.5) | 0.436 | 0.008 |
| gemma-2b | 0.544 [-0.748, 1.836] | 0.83 (152.2) | 0.134 | 0.410 |
| gemma-7b | 0.593 [-0.743, 1.930] | 0.87 (152.4) | 0.141 | 0.385 |
| Mistral-7B-v0.1 | 0.939 [-0.452, 2.329] | 1.32 (152.7) | 0.214 | 0.188 |
| Olmo-3.1-32B-Instruct | 0.964 [-0.852, 2.781] | 1.04 (151.6) | 0.169 | 0.300 |
| pythia-1b-deduped | -0.190 [-1.593, 1.213] | -0.26 (152.9) | -0.043 | 0.791 |
| Qwen2.5-32B-Instruct | 1.266 [-0.249, 2.781] | 1.64 (153.6) | 0.264 | 0.104 |
| Qwen2.5-7B-Instruct | 0.018 [-1.571, 1.607] | 0.02 (153.5) | 0.004 | 0.982 |

Fig 3 shows these comparisons. A value > 0 means that the second sentence of each minimal pair (1b, d, f, h, 2b, d, f, h in Table 1) has a higher surprisal value than the control sentence (1a, c, e, g, 2a, c, e, g), with the control sentence being identical across all conditions.

[INSERT FIG 3 HERE]

**Fig 3**. Surprisal differences (second part – first part) of each minimal pair per condition.

As expected, the LF-grammatical condition reveals the smallest differences between the two parts of each minimal pair, due to the lack of linguistic violations in any of the parts. If there are differences between C2 and C3, they are higher for C3. This is not in the expected direction, based on what one would anticipate if surprisal in LLMs can be directly linked to syntactic processing [9-11, 14]: It is the pragmatically odd condition that reveals the greater surprisal, not the ungrammatical one. If, as Mahowald et al. (2024) argue, LLMs lack functional competence —that involves world knowledge and pragmatic/social knowledge— but have full formal (morphosyntactic) competence [38], C3 should in fact show no surprisal spike, because the type of linguistic violation featured in this condition would presumably fall outside the model's competence. Contrary to that, we find no surprisal differences between pragmatically odd prompts in comparison the ungrammatical ones.

Recall that C4 features "impossible concepts" that violate lexical semantic requirements, but that syntax is intact. Observing this difference in surprisal when the models are presented with non-target syntax vs. non-target semantics is again informative about the claim that LLMs have acquired formal competence, meaning that they are successful in integrating both syntax and semantics into their representations [19, 38]. While Piantadosi (2023) argues that this integration runs contrary to the idea of the *autonomy of* syntax [19]—which amounts to the claim that there is a separate module for syntax, such that certain selectional rules come from the syntactic component, while different requirements come from the semantic component, as an additional interpretive device[39-40]—, our results suggest that indeed a difference in the source of violation exists and in fact LLMs may be sensitive to it: Despite the great overlap in surprisal values between the two conditions C2 and C4 (Figure 3), at a statistical level all the tested models are significantly more surprised by semantic than syntactic violations (Table 4).

**Table 4. Differences between the C2 (HF-grammatical vs. ungrammatical) and C4 (HF-grammatical- vs. semantically odd) condition, estimated as the regression weights of the C4 condition in a linear mixed model in which the C2 condition is the baseline condition.**

| **LLM** | **sem. odd C4: *b* [95% CI]** | ***t (df)*** | **Cohen's *d*** | ***p*** |
|---|---|---|---|---|
| DeepSeek-R1-Distill-Qwen-32B | 4.749 [3.128, 6.369] | 5.74 (152.6) | 0.930 | < .001 |
| gemma-2b | 3.715 [2.433, 4.996] | 5.68 (151.4) | 0.923 | < .001 |
| gemma-7b | 3.650 [2.325, 4.976] | 5.40 (151.5) | 0.877 | < .001 |
| Mistral-7B-v0.1 | 3.778 [2.398, 5.157] | 5.37 (151.9) | 0.871 | < .001 |
| Olmo-3.1-32B-Instruct | 3.771 [1.969, 5.573] | 4.10 (150.7) | 0.668 | < .001 |
| pythia-1b-deduped | 3.349 [1.957, 4.741] | 4.72 (152.1) | 0.765 | < .001 |
| Qwen2.5-32B-Instruct | 3.824 [2.321, 5.327] | 4.99 (152.7) | 0.807 | < .001 |
| Qwen2.5-7B-Instruct | 2.329 [0.753, 3.905] | 2.90 (152.6) | 0.469 | 0.004 |

Last, to determine the effects of frequency and word length, we collected these measures for all critical words (i.e. final nouns), using COCA [34] and syllabifying respectively. Aligning with our previous claims, we find a consistent main effect of condition across models, while frequency differences in critical items do not significantly predict differences in surprisal in any model (all ps $\geq$ .176). Difference in word length, by contrast, shows a more inconsistent influence. It was significantly associated with surprisal differences in Gemma-2B ($\beta = -0.83$, $t = -2.91$, $p = .004$), Gemma-7B ($\beta = -0.58$, $t = -1.97$, $p = .050$), OLMo-3.1-32B-Instruct ($\beta = -1.12$, $t = -2.79$, $p = .006$), and Pythia-1B ($\beta = -0.63$, $t = -2.03$, $p = .043$). The effect was marginal in DeepSeek-R1-Distill-Qwen-32B ($p = .070$), Mistral-7B-v0.1 ($p = .055$), and Qwen2.5-7B-Instruct ($p = .063$), while remaining non-significant in Qwen2.5-32B-Instruct ($p = .205$). Overall, these analyses indicate that although lexical frequency and word length differences contribute to variation in surprisal (which is a result that supports our claim that surprisal should not be viewed as a unique proxy for ungrammaticality or any other type of linguistic violation, as it reflects many other influences such as word frequency, ambiguity, idiomaticity, collocability, semantico-pragmatic soundness, etc.), the persistent positive effect of experimental condition indicates that the observed surprisal differences cannot be reduced to differences in lexical frequency and word length.

# Discussion

Our experiment investigates whether minimal-pair surprisal differences provide a reliable means of identifying the 'grammaticality-ungrammaticality' distinction in LLMs. The answer is negative. While the LLMs we tested are indeed very consistently surprised by violations in the linguistic stimuli, replicating previous studies [9-11], this happens not just for syntactic violations, but also for *other* (namely, semantic) violations and even correct but pragmatically odd language. In fact, the bulk of the highest surprisal values in our experiment were consistently recorded in the semantically odd condition for all models, not the ungrammatical one (Fig 2). We argue that our results bring to the fore two distinct issues, pertaining to (i) the methodology we use to investigate the internalized knowledge of LLMs as well as the status of probabilities as a testing instrument, and (ii) the linguistic knowledge of LLMs and its theoretical and empirical implications.

Starting from (i), finding a fair testing method is a recurrent challenge frequently discussed in the literature [15-16]. While some work has argued that eliciting probabilities is a more appropriate method than direct prompting when it comes to determining the syntactic abilities of LLMs, our results suggest otherwise. Specifically, we argue that probabilities are too blunt a measure to establish a direct link between them and the models' internal syntactic knowledge, contrary to what has been argued in previous literature [9-14, 19]. While recent literature has toned down the strength of this link [24], the association still holds. For instance, Hu et al. (2026) argue that while it is uncontroversial that probability is not the same as grammaticality, the former is a good proxy for obtaining insights for the latter: In their words, string probability can reveal useful information about a probabilistic model's underlying grammatical knowledge [24, p. 135].

Probabilities just indicate whether a sentence is less likely than another —which is the very definition of probability (and hence log probability and surprisal). To illustrate this point with an analogy, using probabilities to determine whether LLMs can tell grammatically possible from impossible prompts is like trying to diagnose PTSD through using life satisfaction questionnaires. When you pair two people starting from the knowledge that one of them has PTSD and the other does not, chances are high that the person with PTSD will indicate less life satisfaction, such that you will find a link between the two. However, if you pair any two people without any prior knowledge about their condition, a lower life satisfaction score alone would not provide a sufficiently specific basis for diagnosing one of them with PTSD. Put differently, it is not that probabilities elicited over minimal pairs necessarily provide a direct measure of the syntactic knowledge of the models (as previous work has argued), but that the minimal-pair experimental design used in previous work artificially restricts possible linguistic violations, allowing one at a time. If only syntactic violations are used in an experiment, the higher surprisal values will inevitably tend to coincide with ungrammaticality. This does not, however, legitimize establishing a unique, one-to-one relation between the two. Overall, observing the great overlap in surprisal values across conditions (Fig 2 and 3), we find it unlikely that binarized, minimal-pair benchmarks can fully capture the intricate complexity of linguistic knowledge [8].

In relation to methodological concerns, when it comes to determining the ability of LLMs to distinguish possible from impossible languages, using the right stimuli is important. As mentioned in the Introduction, Hunter (2024) noted a confound in Kallini et al.'s (2024) prompts, challenging the appropriateness of their experimental design in the context of their key research question (i.e. whether LLMs can learn impossible languages) [21]. We found a second challenge. In the Kallini et al. hierarchy of impossibility, the Shuffle condition (e.g., "his messy books he very . lf he cleans", English equivalent "He cleans his very messy bookshelf") is placed higher than the Reverse condition (e.g., "He cleans his very messy books **R** . lf he", with token R reversing tokens that follow). Our experiment featured only the Reverse condition (Table 1). The reason for not including Shuffle languages in our experimental design has to do with the type of violation that this condition is meant to test. According to Kallini et al., the learnability of a language is more heavily impacted in the Shuffle condition, compared to the Reverse condition, citing the lack of *information locality* in the former. Why does this lack lead to an impossible language? Kallini et al. define Shuffle languages as lacking information locality, an information-theoretic principle suggesting that words that predict each other tend to appear close together in linear order [14]. However, in work that lays the foundations of this principle, the key justification behind it evokes time concerns and memory constraints: "when two elements that predict each other in principle are separated in time, they will not be able to predict each other in practice because by the time the processor gets to the second element, the first one has been partially forgotten. The result is that the second element is less predictable than it could have been, causing excess processing cost" (Futrell 2019: 5) [41]. Thus, while Kallini et al. suggest that Shuffle languages are highly impossible, the question that arises is *for whom*. For humans perhaps yes, but are LLMs bound by the same processing limitations as humans? If not, taking for granted that these languages are as impossible for models as they are for humans is not justified.

LLMs do not possess human-like working memory for manipulating and actively maintaining information during complex tasks like linguistic reasoning. Put another way, LLMs do not possess the *Now-or-Never bottleneck*, a fundamental constraint on human language processing, according to which the human brain parses, compresses, and recodes linguistic input as rapidly as possible, because human memory is fleeting [42]. Models are not bound by such constraints: One can pick up a conversation with an LLM after months, asking the model to summarize the previous response using the same words, and the model will successfully comply with the request. As Goldberg et al. (2025) show in their replication of Dentella et al. (2024), restricting the ability of humans to reread prompts in a linguistic task decreases human accuracy, hence closing the accuracy gap between humans and models that is reported in Dentella et al. [5, 12]. This does not mean the tested models have acquired the relevant linguistic knowledge that will make their performance in a linguistic task more human-like; it means that altering the testing regime in a way that allows human memory constraints to kick in gives a possibly unfair advantage to the models, because they are not bound by the Now-or-Never bottleneck and the processing limitations that derive from it.

Returning to Kallini et al., a more superficial explanation exists as to why LLMs may perform worse in Shuffle languages than in Reverse languages: The more pronounced the disturbances in the form are, the more likely the models are to be surprised, because the prompt they encounter offers a very weak match to the forms they have seen in the training data. This does not entail that the models are sensitive to the underlying cause of the disturbances (e.g., lack of information locality, illicit count-based dependencies, non-canonical word orders, etc.). Telling in this respect are the results of a recent leet (l33t) decoding experiment [43]: Models perform better in decoding prompts with two letter-number substitutions (e.g., "C8t0 orang8tan is availabl0 for sal0" for "Cute orangutan is available for sale") than three (e.g., "3e resigned 3is job wi23o92 any warning" for "He resigned his job without any warning"). Humans performed equally well and almost at ceiling in both conditions. More importantly, all models gave some non-sensical answers (e.g., "Eight eighteen resigned seventeen this eight twenty-one any ten warning" for "3e resigned 3is job wi23o92 any warning"). Such answers were totally absent from human responses. If high surprisal upon encountering form disturbances meant that LLMs have internalized the generalizations that govern the use of the words in this prompt, this answer should not have emerged. To sum up, we argue that probabilities cannot be interpreted as proxies for grammatical well-formedness. Consequently, it remains to be demonstrated whether models are capable of drawing finer distinctions, within the realm of impossible, of what defies syntax vs. semantics vs. pragmatics.

At the same time, we also argue that the other side of the debate regarding the ability of LLMs to tell possible from impossible languages should also re-evaluate and possibly better qualify its position. This brings us to point (ii): the internalized linguistic knowledge of LLMs and its theoretical implications. We found that models *can* tell apart well-formed from ill-formed prompts to a good degree, assigning lower probabilities to the latter (Fig 2). Kallini et al. discuss Chomsky et al. (2023) who argue in a popular-press article that "ChatGPT and similar programs are, by design, unlimited in what they can "learn" (which is to say, memorize); they are incapable of distinguishing the possible from the impossible. […] Whereas humans are limited in the kinds of explanations we can rationally conjecture, machine learning systems can learn both that the earth is flat and that the earth is round. They trade merely in probabilities that change over time." [17] This claim by Chomsky et al. is both on the right track and in need of reconsideration. It is on the right track in the sense that LLMs form statistical correlations that are constantly shifting and oscillating [4-6]. Work on the so-called AI-bewitchment has highlighted the LLM lack of the *constancy* needed to track reference points and contradictions throughout a dialogue [44].

Yet these statistical correlations inevitably capture some aspects of the training data and, by extension, of the real world; these are strings of text with *natural histories* [45]. This does not mean that LLMs grasp or refer to these histories, but that their outputs can be *about* them [46]. Put differently, there are glimpses of real-world conditions in the LLM outputs, that humans —not LLMs— can recognize as familiar or not, meaning that we can assess their soundness and truth validity based on our real-world knowledge [46]. In this sense, LLMs are not unlimited in what they can "learn", as Chomsky et al. (2023)

argue: They only "learn", in a non-deterministic way, forms that are in their training data [47]. They are not fully incapable of distinguishing the possible from the impossible, if by "impossible" we mean a pattern that is never attested in the training data. LLMs are surprised by such patterns: Our results suggest that semantically odd phrases such as 'drinkable cold laptops' (Table 1) are consistently assigned low probabilities, showing that the models are indeed sensitive to this violation. This does not mean that the models have human-like semantics that enables them to *understand* the nature of the violation, only that they can recognize in a mechanistic way —akin to Searle's Chinese room thought experiment [48]— that the form they receive is a very weak match to the forms they have seen in their training, hence the surprise.

If by "incapable of distinguishing the possible from the impossible", Chomsky et al. mean that, upon receiving training data that feature "impossible concepts", the models may learn to use them, this is of course true, but it is a trivial point, also true for humans who can invent and learn a lexical label for any "impossible concept"— any attempt to provide a counterexample illustrates this point. As for impossible grammars, there is no solid evidence of what exactly is unlearnable in humans and *why*, meaning whether we are dealing with purely syntactic constraints or general principles of efficient communication. Instead, the evidence for humans has been much narrower: When the human brain computes impossible structures, the language networks are progressively inhibited [22]. While it is true that count-based dependencies are not attested in any natural language [18], we still lack a solid theory of what is impossible in human grammar. Piantadosi (2023) is right to notice that the idea of "impossible languages" has frequently been evoked but has never been empirically justified [19]. Kallini et al. construe "impossible" through prompts that are blatantly violating many lexical and syntactic rules simultaneously (e.g., "messy books his he very . If He cleans") [14]. The result is that their hierarchy of impossible languages is arbitrary, because it does not come with a set of explanations that quantify the different degrees of separation for the impossible languages, based on either linguistic rules or levels of linguistic analysis (e.g., language X is higher on the hierarchy than language Y because it violates more rules). Word salads of that sort simply do not allow us to tease apart the precise abilities of LLMs in order to understand how they fare in terms of semantics, morphology, syntax, or pragmatics.

This point has direct implications for our ability to assess the internal knowledge of LLMs. Kallini et al. comment on how previous research provides only a few concrete examples that involve counting word positions to mark features like negation and agreement, noting that this is a narrow definition of what is impossible [14, 22]. The problem is that Kallini et al. broaden the scope of what is impossible to the point that any principled explanation of what this impossibility boils down to is undiscernible. In their prompt "messy books his he very . lf  He cleans", does the impossibility stem from impaired syntax, non-target semantics, or infelicitous pragmatics? Is there any linguistic notion (e.g., meaning, reference, number of syntactic rules violated) that can serve as a criterion to decide whether "messy books his he very . lf  He cleans" is more impossible than "lf he R books messy very his cleans He", as their hierarchy predicts? Would these prompts give meaningful results if tested with humans in order to establish accurate baselines? In the absence of human comparisons, it is hard to understand whether LLMs filter out as impossible the same prompts that humans do.

All in all, it remains to be established whether a hierarchy of grammatical impossibility can be construed at all. At present, it is not clear that it can, because nobody knows what the space of impossible languages is and what it involves [19]. In humans, ungrammaticality is not a matter of degree [49]. This means that either a structure is attested in some natural languages, or not. Similarly, a prompt is either grammatical or ungrammatical; it contains a violation of at least one rule of grammar, or it does not. In principle, an ungrammatical prompt cannot be more ungrammatical than another ungrammatical prompt, and a structure that is never attested in any natural language cannot be more unattested than another structure that is also never attested in any natural language. This challenges the ordering of impossible languages in Kallini et al. on theoretical grounds: Constructing hierarchies of syntactic impossibility is not informative if there is no prior justification of *how* (i.e. on what grounds) something impossible can be judged as more impossible/unattested than something else (cf. Figure 1). In science, the concept of measurement relies on a collective agreement over the use of a valid metric. As the field is still developing the right metrics for determining the linguistic knowledge of LLMs, and in the absence of a metric of grammatical impossibility, we argue that any strong claims regarding models being (in)sensitive to what crosses the limits of the impossible are premature. Recording the presence and magnitude of such errors is important for, echoing Piantadosi (2023), one must be frank about the state of the art for models that seek to capture human language [19].

A limitation of the present study concerns the possibility of cross-linguistic transfer. For instance, the C2 condition was designed to introduce an ungrammatical word-order pattern in English, but the reversed adjective+noun order used in our stimuli is grammatical in a number of other languages (e.g., Spanish and other Romance languages). Since the models tested here are multilingual, we cannot exclude the possibility that exposure to such patterns in other languages may contribute to their probability assignments for C2 prompts. More generally, when using multilingual models to investigate language-specific syntactic constraints, it is difficult to establish that a given violation is entirely independent of patterns that may occur in other languages represented in the training data. Thus, the present results should be interpreted with this possibility in mind. Future studies could address this issue by testing syntactic properties whose relevant violations are more clearly language-specific and by systematically controlling for possible cross-linguistic transfer.

A further limitation concerns the scope of the linguistic phenomena represented in our benchmark. Although the experiment includes three distinct types of violations — syntactic, semantic, and pragmatic— and matches these conditions in terms of the number of prompts tested, each condition instantiates only a single type of linguistic deviation. Thus, our results should not be taken to establish that probability-based measures behave in the same way across all possible forms of syntactic, semantic, or pragmatic deviation. In particular, the syntactic condition tested here represents one specific type of word-order violation, and other syntactic phenomena may yield different patterns of surprisal. A broader assessment would therefore require extending the benchmark to a wider range of phenomena within each domain. Nevertheless, the present design provides a controlled comparison across three distinct sources of deviation, allowing us to test whether increased surprisal is uniquely associated with syntactic ungrammaticality in this benchmark. More broadly, our results challenge the use of probability assignment as a general diagnostic of grammaticality, at least for the range of syntactic, semantic, and pragmatic phenomena represented in our benchmark

# Outlook

We provided a collection of data from different experiments pulling into question previous proposals that LLMs, at their current stage of development, have internal representations of (un)grammaticality. Probability assignment is not a reliable proxy for grammaticality, hence any conclusions based on this method need further corroboration through a different testing method. Overall, while many scholars talk about LLMs as a success story in capturing language —as a hallmark characteristic of human cognition—, it seems that such claims should be taken as future goals rather than actual achievements. This is particularly important in the context of weaponizing LLMs as a success story in the battle of linguistic frameworks. While Piantadosi (2023) argues that this unmatched success of LLMs can be viewed as humiliating for a subfield of linguistics (i.e. generative linguistics) that, according to him, has spent decades deriding these tools [19], it seems that the successes, which are indeed real, are at times portrayed as more pronounced than they really are. This contributes to the so-called *AI-bewitchment*, in the sense that overhyped and exaggerated capabilities contribute to the popular misconception that LLMs meet or exceed human-level baselines, leading to belief distortion [50]. However, we agree with the conclusion that "now we can do better" [19]. This entails recording both successes and failures. To this end, our experiment has provided results suggesting that both sides of the debate should reconsider their position in the face of new evidence: LLMs can tell apart possible from impossible prompts to some degree, but at the same time, this does not constitute reliable evidence that they have internalized the relevant syntactic rules, just that they are surprised when they see patterns that they have not encountered in their training phase. It remains to be demonstrated whether in the future LLMs can develop a human-like grasp of impossible language, and this is only possible through rigorous experiments, adequate comparisons with humans, and complementary methodologies such as controlled syntactic evaluations, probing techniques, and representation analyses.

# References

1. Savage, N. 2024. Beyond Turing: Testing LLMs for intelligence. *Communications of the ACM* 67(9), 10-12.
2. Jones, C. R. & Bergen, B. K. 2025. Large Language Models pass the Turing Test. https://arxiv.org/abs/2503.23674
3. Barattieri di San Pietro, C., Frau, F., Mangiaterra, V., Bambini, V. 2023. The pragmatic profile of ChatGPT: Assessing the communicative skills of a conversational agent. *Sistemi Intelligenti* 35(2), 379-399.

4. Dentella, V. F. Günther & E. Leivada. 2023. Systematic testing of three Language Models reveals low language accuracy, absence of response stability, and a yes-response bias. *PNAS* 120, e2309583120.
5. Dentella, V. F. Günther, E. Murphy, G. Marcus & E. Leivada. 2024. Testing AI on language comprehension tasks reveals insensitivity to underlying meaning. *Scientific Reports* 14, 28083.
6. Dentella, V. F. Günther & E. Leivada. 2025. Language learning in vivo vs. in silico: Size matters but Larger Language Models still do not comprehend language on a par with humans due to impenetrable semantic reference. *PLoS ONE* 20(7), e0327794.
7. Zhou, H., Y. Hou, Z. Li, X. Wang, Z. Wang, X. Duan & M. Zhang. 2023. How well do Large Language Models understand syntax? An evaluation by asking natural language questions. arXiv:2311.08287
8. Vázquez Martínez, H. J., A. Heuser, C. Yang & J. Kodner. 2023. Evaluating Neural Language Models as cognitive models of language acquisition. Proceedings of the 1st GenBench Workshop on (Benchmarking) Generalisation in NLP, 48–64.
9. Hu, J. & R. Levy. 2023. Prompting is not a substitute for probability measurements in large language models. In Proceedings of the 2023 Conference on Empirical Methods in Natural Language Processing, 5040–60.
10. Hu, J., & M. C. Frank. 2024. Auxiliary task demands mask the capabilities of smaller language models. Proceedings of the 1st Conference on Language Modeling.
11. Hu, J., K. Mahowald, G. Lupyan, A. Ivanova & R. Levy. 2024. Language models align with human judgments on key grammatical constructions. PNAS 121(36), e2400917121.
12. Goldberg, A. E., S. Rakshit, J. Hu & K. Mahowald. 2026. A suite of LMs comprehend puzzle statements as well or better than humans *Open Mind 10*, 431–440.
13. Gulordava K, Bojanowski P, Grave E, Linzen T, Baroni M. 2018. Colorless green recurrent networks dream hierarchically. In *Proceedings of the 2018 Conference of the North American Chapter of the Association for Computational Linguistics*, 1195–205.

14. Kallini, J., I. Papadimitriou, R. Futrell, K. Mahowald & C. Potts. 2024. Mission: Impossible Language Models. Proceedings of the 62nd Annual Meeting of the Association for Computational Linguistics (Volume 1: Long Papers), 14691–14714.
15. Leivada E, Dentella V, Günther F. 2024a. Evaluating the language abilities of Large Language Models vs. humans: Three caveats. *Biolinguistics* 18, e14391.
16. Ivanova, A. A. 2025. How to evaluate the cognitive abilities of LLMs. Nature Human Behaviour 9, 230–233.
17. Chomsky, N. I. Roberts & J. Watumull. 2023. Noam Chomsky: The false promise of Chat-GPT. The New York Times.
18. Moro, A., M. Greco & S. F. Cappa. 2023. Large languages, impossible languages and human brains. *Cortex* 167, 82-85.
19. Piantadosi, S. T. 2023. Modern language models refute Chomsky's approach to language. In: Gibson E, Poliak M. From fieldwork to linguistic theory: A tribute to Dan Everett. Berlin: Language Science Press, 353-414.
20. Adger, D. (2018). The autonomy of syntax. In N. Hornstein, H. Lasnik, P. Patel-Grosz & C. Yang, Syntactic Structures after 60 years. The Impact of the Chomskyan Revolution in Linguistics, 153-175. De Gruyter Mouton.
21. Hunter, T. 2025. Kallini et al. (2024) do not compare impossible languages with constituency-based ones. Computational Linguistics 51 (2), 641–650.
22. Musso, M, Moro A, Glauche V, Rijntjes M, Reichenbach J, Büchel C & Weiler C. 2003. Broca's area and the language instinct. *Nature Neuroscience* 6(7): 774-781.
23. Yang, X., T. Aoyama, Y. Yao & E. Wilcox. 2025. Anything goes? A crosslinguistic study of (im)possible language learning in LMs. Proceedings of the 63rd Annual Meeting of the Association for Computational Linguistics (Volume 1: Long Papers), 26058–26077.
24. Hu, J., E. Wilcox, S. Song, K. Mahowald, R. Levy. 2026. What can string probabilities tell us about grammaticality? *Transactions of the Association for Computational Linguistics 14*, 124–146.
25. Kauf, C., A. A. Ivanova, G. Rambelli, E. Chersoni, J. S. She, Z. Chowdhury, E. Fedorenko & A. Lenci 2024. Event knowledge in Large Language Models: The gap Between the impossible and the unlikely. Cognitive Science 47, e13386.
26. Gemma Team et al. 2024. Gemma: Open Models Based on Gemini Research and Technology. https://arxiv.org/abs/2403.08295.

27. Biderman, S. et al. 2023. Pythia: A Suite for Analyzing Large Language Models Across Training and Scaling. https://arxiv.org/abs/2304.01373.
28. Jiang, A. Q. et al. 2023. Mistral 7B. https://arxiv.org/abs/2310.06825.
29. DeepSeek-AI et al. 2025. DeepSeek-R1: Incentivizing Reasoning Capability in LLMs via Reinforcement Learning. https://arxiv.org/abs/2501.12948.
30. Team Olmo et al. 2025. Olmo 3. https://arxiv.org/abs/2512.13961.
31. An Yang et al. 2024. Qwen2 Technical Report. https://arxiv.org/abs/2407.10671
32. Hale, J. 2001. A probabilistic Earley parser as a psycholinguistic model. In Proceedings of NAACL, Vol. 2, 159–166.
33. Levy, R. 2008. Expectation-based syntactic comprehension. Cognition 106(3): 1126–1177.
34. Davies, M. 2008-. *The Corpus of Contemporary American English (COCA)*. Available online at https://www.english-corpora.org/coca/.
35. Bates, D., Maechler, M., Bolker, B. & Walker, S. 2015. Fitting linear mixed-effects models using lme4. *J. Stat. Softw.* 67 (1), 1–48.
36. Kuznetsova, A., Brockhoff, P. B., & Christensen, R. H. 2017. lmerTest package: tests in linear mixed effects models. *J. Stat. Softw.*, 82, 1-26.
37. Leivada, E., Günther, F., & Dentella, V. 2024b. Reply to Hu et al: Applying different evaluation standards to humans vs. Large Language Models overestimates AI performance. PNAS 121 (36) e2406752121.
38. Mahowald K, Ivanova AA, Blank IA, Kanwisher N, Tenenbaum JB, Fedorenko E. 2024. Dissociating language and thought in large language models. Trends in Cognitive Sciences 28(6): 517–540.
39. Chomsky, N. 1957. *Syntactic Structures*. Mouton.
40. Chomsky, N. 1965. *Aspects of the Theory of Syntax*. MIT Press.
41. Futrell, R. 2019. Information-theoretic locality properties of natural language. In Proceedings of the First Workshop on Quantitative Syntax (Quasy, SyntaxFest 2019), 2–15.
42. Christiansen MH & Chater N. 2016. The Now-or-Never bottleneck: A fundamental constraint on language. *Behavioral and Brain Sciences* 39: e62.
43. Leivada, E., G. Marcus, F. Günther & E. Murphy. 2026. A sentence is worth a thousand pictures: Can Large Language Models understand hum4n l4ngu4ge and the w0rld behind w0rds? *Philos Trans A Math Phys Eng Sci* 14, 384 (2320): 2025008.

44. Bottazzi Grifoni, E. & R. Ferrario. 2025. The bewitching AI: The illusion of communication with Large Language Models. Philosophy & Technology 38: 61.
45. Mandelkern, M. & T. Linzen. 2024. Do Language Models' words refer?. *Computational Linguistics* 50 (3): 1191–1200.
46. Baggio, G. & E. Murphy. 2024. On the referential capacity of language models: An internalist rejoinder to Mandelkern & Linzen. arXiv:2406.00159
47. Bender, E. M., Gebru, T., McMillan-Major, A. & Shmitchell, S. 2021. On the dangers of stochastic parrots: Can Language Models be too big? *Proceedings of the 2021 ACM Conference on Fairness, Accountability, and Transparency*, 610–623.
48. Searle, J. R. 1980. Minds, brains, and programs. *Behavioral and Brain Sciences* 3(3), 417-424.
49. Leivada, E., & Westergaard, M. 2020. Acceptable ungrammatical sentences, unacceptable grammatical sentences, and the role of the cognitive parser. Frontiers in Psychology 11, 364.
50. Kidd, C. & A. Birhane. 2023. How AI can distort human beliefs. *Science* 380 (6651), 1222–1223.